\documentclass[letterpaper]{article} 
\usepackage[]{aaai23}  
\usepackage{times}  
\usepackage{helvet}  
\usepackage{courier}  
\usepackage[hyphens]{url}  
\usepackage{graphicx} 
\usepackage{natbib}  
\usepackage{caption} 
\usepackage{algorithm}
\usepackage{algorithmic}

\usepackage{newfloat}
\usepackage{listings}
\DeclareCaptionStyle{ruled}{labelfont=normalfont,labelsep=colon,strut=off} 
\floatstyle{ruled}
\newfloat{listing}{tb}{lst}{}
\floatname{listing}{Listing}
\usepackage{tabularx}
\usepackage{booktabs}
\usepackage{tikz}
\usetikzlibrary{matrix, positioning, decorations.pathreplacing}
\tikzset{brace/.style={decorate, decoration={brace}},
  brace mirrored/.style={decorate, decoration={brace,mirror}},
}

\usepackage{xcolor}
\definecolor{RA}{HTML}{639B30}
\definecolor{RB}{HTML}{B8AC2B}
\definecolor{RC}{HTML}{F8B830}
\definecolor{RD}{HTML}{EF7D29}
\definecolor{RE}{HTML}{E52421}

\title{Energy Efficiency Considerations for Popular AI Benchmarks}
\author{
    Raphael Fischer,
    Matthias Jakobs,
    Katharina Morik
}
\affiliations{
    TU Dortmund University, AI Group, 44227 Dortmund, Germany\\
    firstname.lastname@tu-dortmund.de\\
    
}

\begin{document}

\maketitle

\begin{abstract}    
Advances in artificial intelligence need to become more resource-aware and sustainable.
This requires clear assessment and reporting of energy efficiency trade-offs, like sacrificing fast running time for higher predictive performance.
While first methods for investigating efficiency have been proposed, we still lack comprehensive results for popular methods and data sets.
In this work, we attempt to fill this information gap by providing empiric insights for popular AI benchmarks, with a total of 100 experiments.
Our findings are evidence of how different data sets all have their own efficiency landscape, and show that methods can be more or less likely to act efficiently.

\end{abstract}

\section{Introduction}

Despite recent calls for making artificial intelligence (AI) more sustainable \cite{Strubell/etal/2020a}, many works still prioritize predictive quality over resource awareness.
As a result, innovation in the field is for the most part accompanied by tremendous computational efforts \cite{Schwartz/etal/2020b}.
In deep machine learning (ML), resource-heavy training is often necessary to obtain highly accurate models with reasonably small energy footprints \cite{DBLP:journals/corr/abs-1905-11946,DBLP:journals/corr/HowardZCKWWAA17}.
While explicitly reporting and comparing such efficiency trade-offs is crucial, respective works so far mostly investigated singular application scenarios \cite{DBLP:journals/corr/abs-2104-10350,Martin/etal/2019a,sogood2022}.
Even for popular benchmark data sets, it is hard to find concise information on energy draw and efficiency.

In this paper, we contribute to ML sustainability research by extending upon recently published work for assessing efficiency \cite{sogood2022}.
Parts of the methodology are inspired by the EU's well-established energy label system \cite{eu-washingmachines}, and allow for a similarly innovative form of communicating such information for AI.
Making intricate model properties comprehensible to non-experts is a difficult but yet important step towards trustworthy AI \cite{yeswecare,Brundage/etal/2020a}.
While the original paper investigated deep computer vision models for ImageNet classification, we successfully applied their methodology to a wider range of benchmark data and methods from \texttt{scikit-learn} \cite{scikit-learn} and \texttt{OpenML} \cite{vanschoren2014openml}.
These new applications required us to adapt and further enhance the existing framework implementation, all code is (preliminarily) available at \url{https://github.com/raphischer/sklearn-energy-efficiency}.
As benchmarks are popular for proof-of-concept testing in science and industry, we hope that our investigation will make the assemblage of future AI systems more resource-aware.

\section{Investigating Energy Efficiency\\of Machine Learning}
We start off by explaining the necessary methodology for assessing the energy efficiency of ML tasks.
It is largely based on the work of \citeauthor{sogood2022}, where it is also explained in more technical detail.
We here resort to describing key concepts, and explain how we extend them for this work.

\subsection{Characterizing ML Experiments}

Due to the vastly different use cases for AI, quantifying efficiency is non-trivial.
\citeauthor{sogood2022} therefore characterized any ML experiment $X$ that shall be tested for efficiency to consist of a \emph{configuration} and \emph{environment}.
The former combines information on the task to solve (e.g., inference, training, robustness testing, $\dots$), as well as the data set and model (alongside all hyperparameters).
The environment is specified by the hardware and software used for experiment execution.
Carbon emissions could be included by making the local energy mix (which is assumed to be constant) part of the environment \cite{DBLP:journals/corr/abs-2104-10350}.

The relevant properties for assessing how efficiently a task is solved are called \emph{metrics} (e.g., accuracy, model size, power draw).
The experiment configuration determines the applicable metrics $m_i\in M_X$.
Naturally, running experiments with different configurations allows for comparing their efficiency:
For a fixed task, data set, and environment, one can investigate the performance of varying models.
It might also be interesting to examine the efficiency of a particular model when deployed in different environments.
Note that configurations and environments may also be infeasible.
To give an example, the choice of data usually limits the applicable models, and certain models might also require specific soft- or hardware.

\subsection{Calculating Model Efficiency}
\label{subsec:efficiency assessment}

The energy efficiency of a given experiment $X$ can be assessed via a three-step procedure.
First, numeric values $\mu_i(X)$ for each metric $m_i\in M_X$ are aggregated from the experiment log files.
This is straightforward for some metrics (e.g., accuracy), but others, like power draw, require specialized hardware-dependent software tools \cite{henderson2020systematic}.

Afterwards, the metric values are projected onto an \emph{index scale} by comparing them against a reasonable, application-specific reference model $X^*$.
By using such a relative scale, the numeric values of different metrics become comparable:
\begin{equation}\label{eq:index}
    \iota_i(X^\circ)=\left(\frac{\mu_i(X)}{\mu_i(X^*)}\right)^{\sigma_i}
\end{equation}
For the reference (i.e., $X = X^*)$, the index scores of all metrics are one, while other models are assessed in relation to the reference.
For intuitive comparability, higher index values should always indicate improvement, which is achieved by the constant $\sigma_i$.
It acts as an indicator of whether higher values ($\sigma_i=+1$, e.g., accuracy) or lower values ($\sigma_i=-1$, e.g., power draw) are preferable for the $i$-th metric.
As an example, a model with $20\%$ higher accuracy than the reference receives an index score of $1.2$, while a running time index of $0.9$ indicates $10\%$ slower execution.
Note that the reference performance needs to be individually measured for each environment in order to obtain sound index values.

For making efficiency information more easily comprehensible, each index scale is divided into five \emph{rating bins}, such that each value is assigned an \colorbox{RA}{A}-\colorbox{RE}{E} rating.
The individual metric ratings can then be summarized into a \emph{compound model rating} via a (weighted) median.
Weighting metrics allows to lower the impact of highly correlated metrics (e.g., different accuracy scores), or fine-tune the efficiency assessment to a specific use case (for example when facing tight memory restrictions).
Similar to the EU energy label system, the application-specific reference model, weights, and rating bins can be determined by experts \cite{eu-washingmachines}.
Periodic updates of the rating system to novel technological advances can be easily performed by choosing new references.
Additionally, \citeauthor{sogood2022} showed how hybrid energy labels can convey efficiency information in a more comprehensible way, bridging the gap between AI experts and less informed target groups.

\subsection{Extensions}

\begin{table}
    \begin{center}
        \includegraphics[]{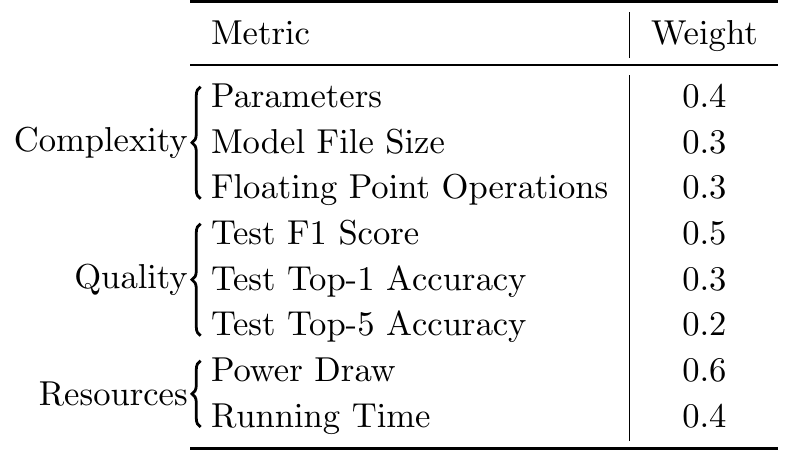}
    \end{center}
    \caption{Categories of efficiency metrics, along with their associated weight for determining the compound rating.}
    \label{tab:metric_weights}
\end{table}

So far, energy efficiency was only investigated for ImageNet classification with deep learning \cite{sogood2022}.
The proposed experiment characterization was therefore restricted to models, where here we discuss much more diverse ML \emph{methods}, such as $k$-nearest neighbors (kNN), support vector machines (SVM), Gaussian naive Bayes (GNB), logistic and ridge regressions (LR, RR), stochastic gradient descent variants of linear models (SGD), adaptive boosting ensembles (AB), (extra) random forests (RF and XRF), and simple multilayer perceptrons (MLP) \cite{scikit-learn}.
Note that this change in terminology does not really alter the proposed methodology, as utilizing these methods also results in specific models.

We investigated the same metrics as \citeauthor{sogood2022}, except for the per epoch training metrics, which are not applicable for most of our methods.
Additionally, we calculate and report the $F_1$ score.
As some models do not provide individual class probabilities, we in these cases resort to reporting the top-1 accuracy twice (since top-1 is a tight lower bound of top-5 accuracy).
As an extension to the original work, we propose to explicitly categorize metrics as displayed in Table \ref{tab:metric_weights}.
Our reasoning is that every ML experiment will have some form of inherent complexity, quality, and resource demand, captured by individual sets of metrics.
When assigning weights, we decided to make each group equally important for the final rating (i.e. weights within each group sum up to one).
Note how power draw received the highest weight to prioritize energy efficiency.

\section{Results and Evaluation}

\begin{table*}[t]
    \begin{center}
        \includegraphics[]{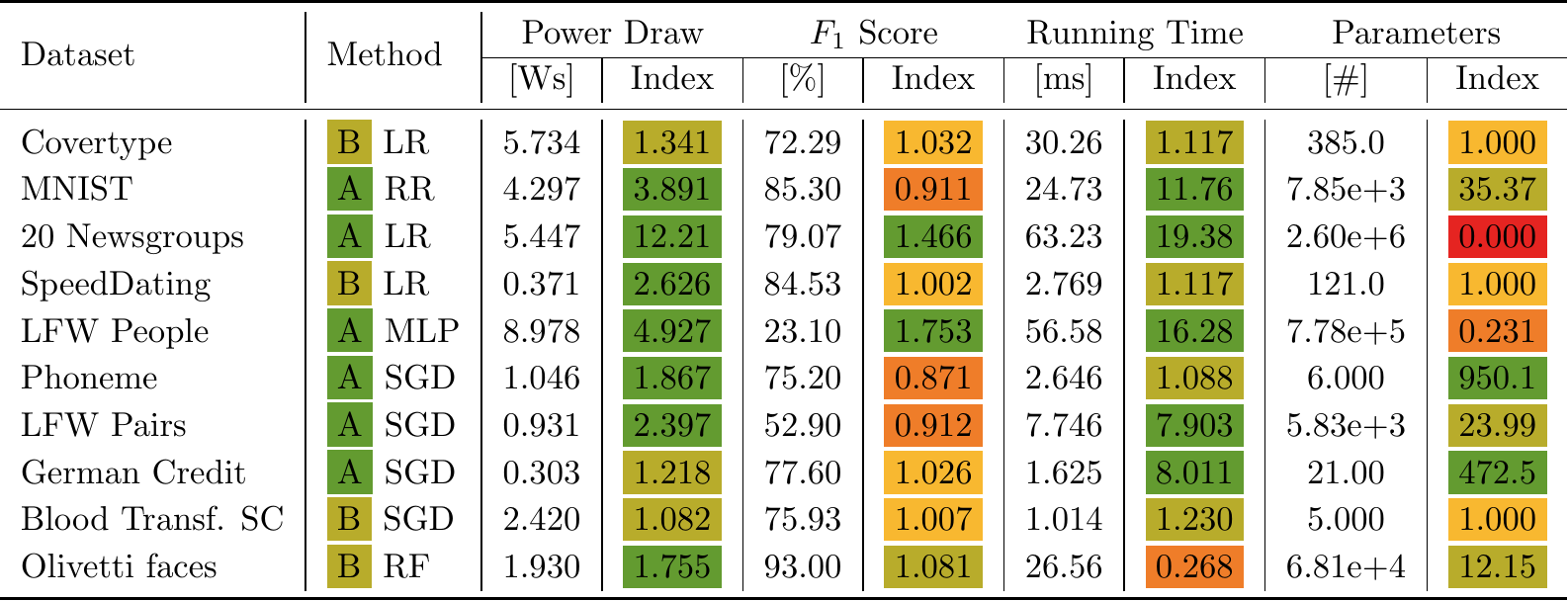}
    \end{center}
    \caption{Most efficient ML method for each data set, along with the individual metric values and index scores. Colors indicate the metric or compound model rating. If the best compound rating was assigned to multiple models, we selected the one with best power draw.}
    \label{tab:best_models}
\end{table*}

All aforementioned methods were deployed on 10 popular classification data sets from varying domains and formats, including tabular, images, and text (full list in Table \ref{tab:best_models}).
First, we performed a random search for finding optimal application-specific hyperparameters.
It tested 50 random assignments for each method $\times$ data combination in a $5$-fold cross-validation on $75\%$ of the data.
This split was also used for training the final model, while the remaining $25\%$ were used as test data.
We estimated power draw with the help of \texttt{RAPL} and \texttt{psutil}, while floating point operations were profiled via \texttt{PAPI}.
An Intel Xeon W-2155 PC with \texttt{Python 3.8} and \texttt{scikit-learn 1.1.2} was used as execution environment.
The application-specific reference methods were chosen to be as mediocre as possible, such that we obtain reasonably worse and better options.
In the following, we discuss the efficiency of inference on the test split, and recommend to take a deeper dive with our interactive webpage\footnote{\url{https://github.com/raphischer/sklearn-energy-efficiency}}.

Firstly, Table \ref{tab:best_models} lists the performance of the best (i.e., most efficient) methods on all benchmarks, sorted by data set size.
We chose the methods with best compound rating, prioritized by power draw when facing multiple candidates.
We here display the compound rating as well as the highest-weighted metrics, along with index scoring and respective ratings (indicated by color).
As expected, there is no ``master of all'', and each method trades certain aspects of efficiency.
As an example, we see that on Covertype data, the LR's low power draw and running time justifies the mediocre scores on other metrics.

\begin{figure}
\centering
    \includegraphics[width=\linewidth]{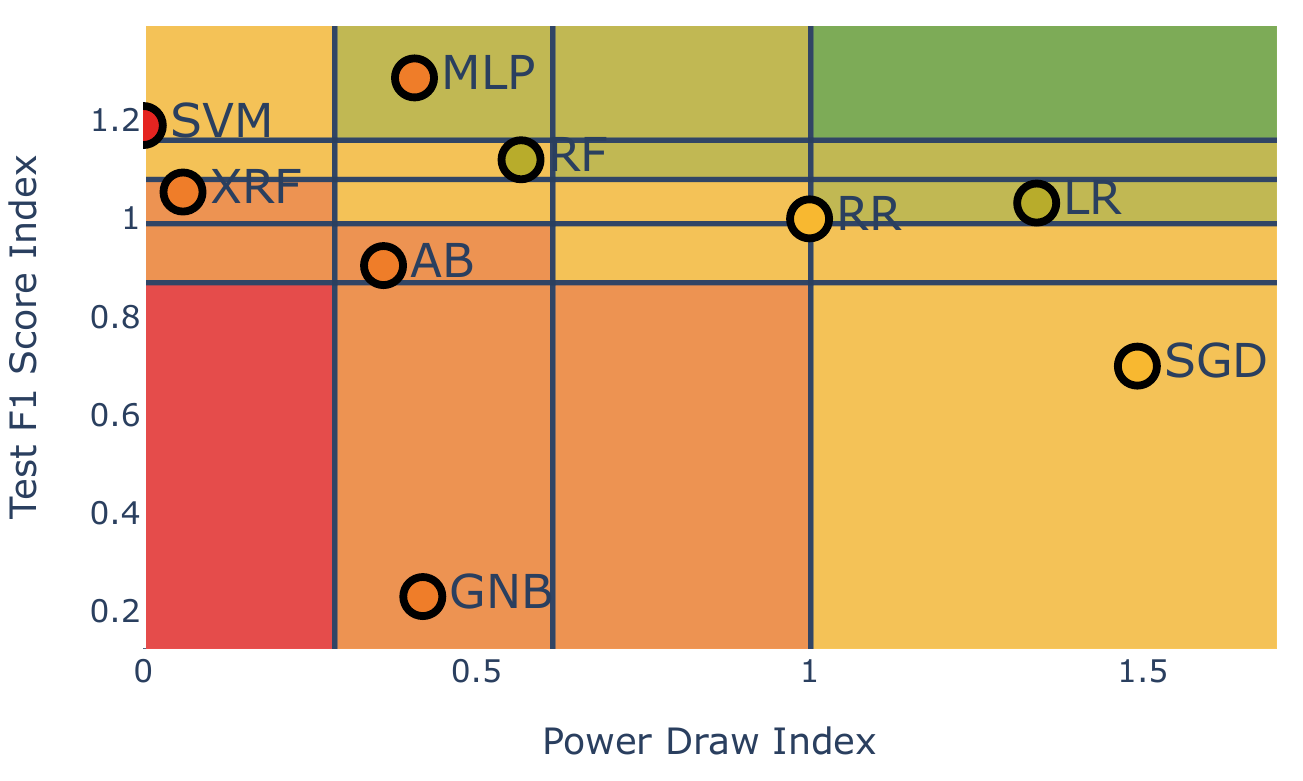}
    \caption{Scatter plot of most important model metrics (power draw versus $F_1$ score). A higher value represents better performance. Model colors indicate the compound rating.}
    \label{fig:scatter-tradeoff}
\end{figure}

\begin{figure}
\centering
    \includegraphics[width=.31\linewidth]{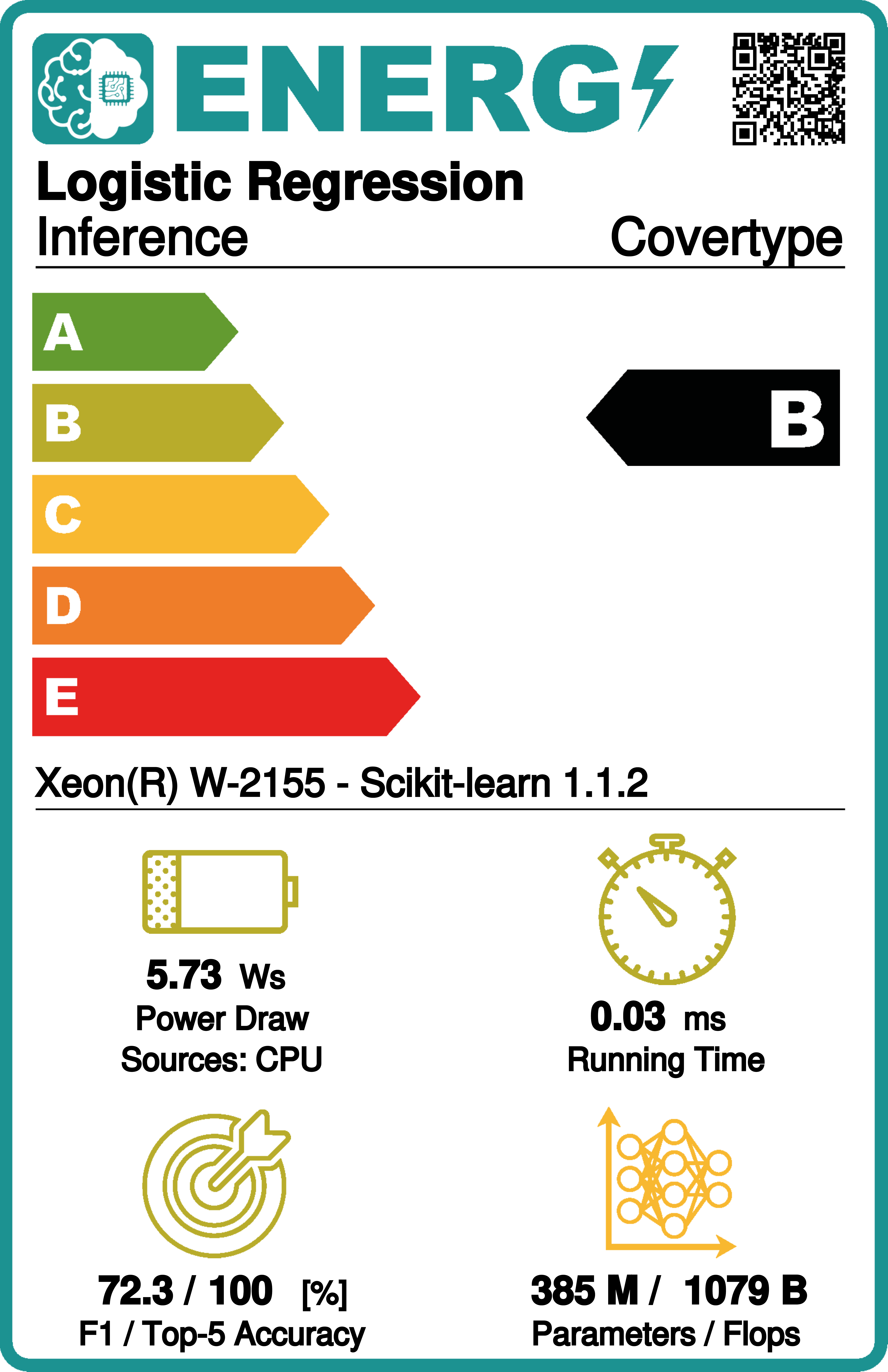}
    \hfill
    \includegraphics[width=.31\linewidth]{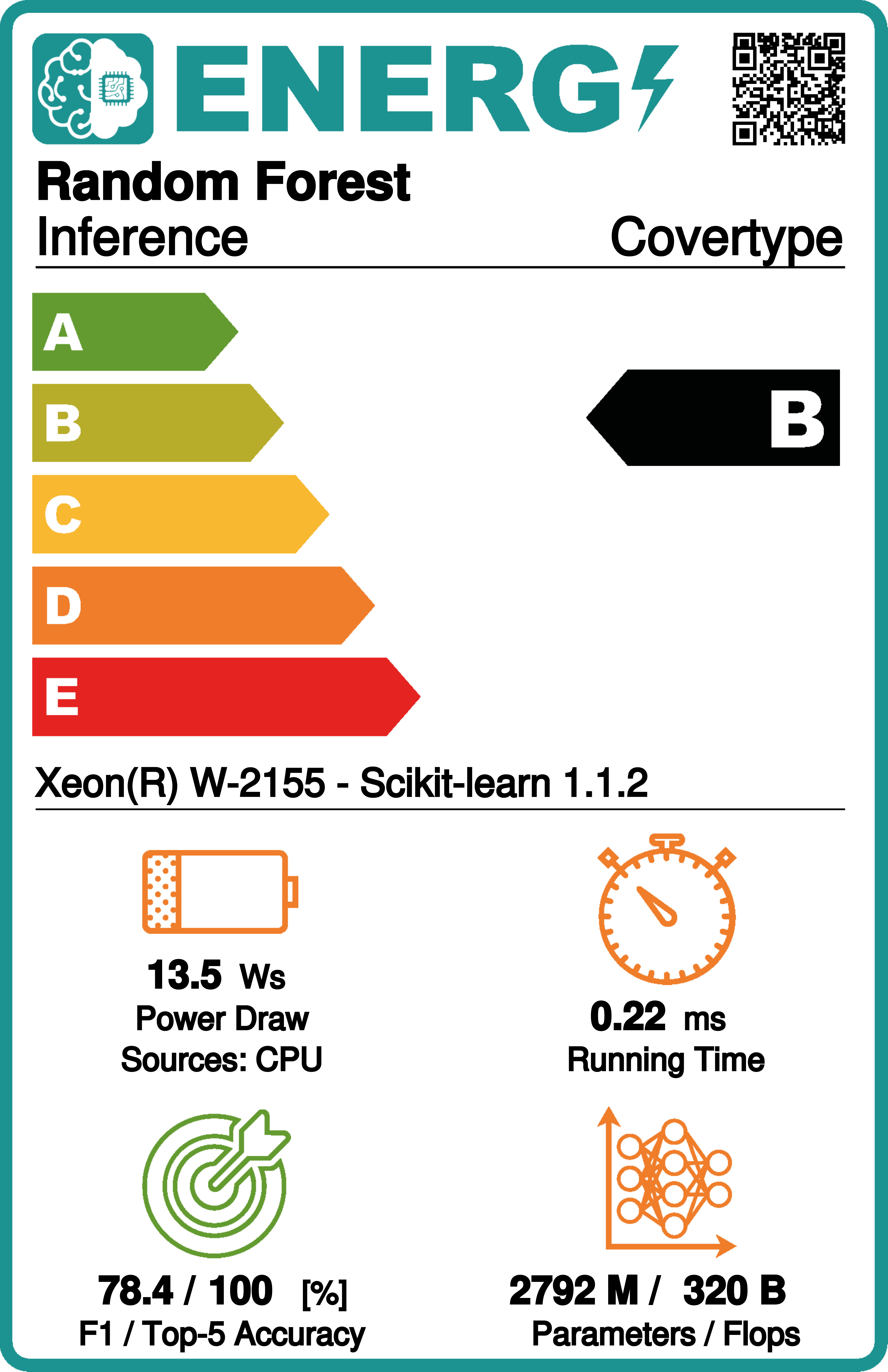}
    \hfill
    \includegraphics[width=.31\linewidth]{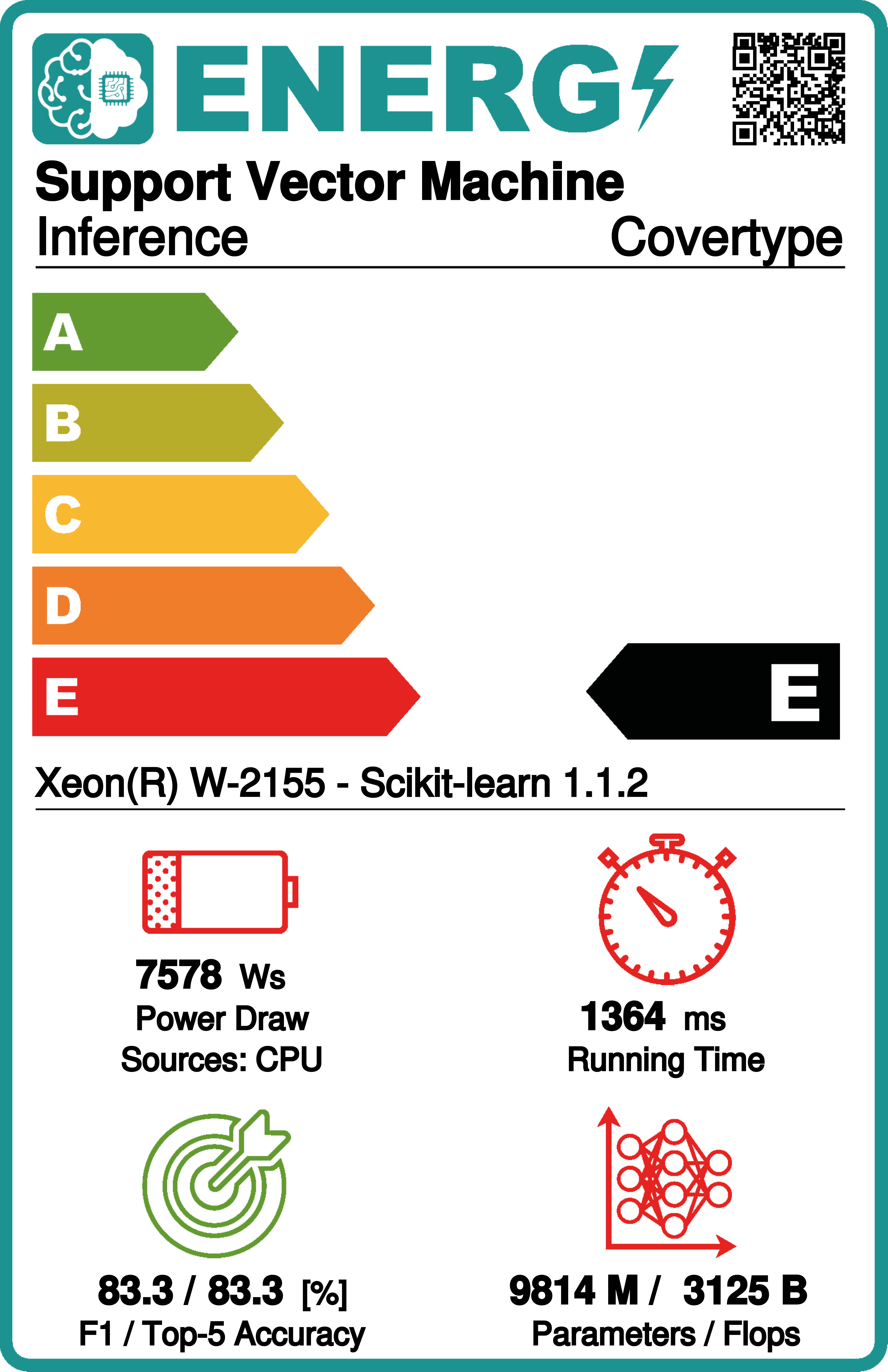}
    \caption{Exemplary energy labels for the most and least efficient methods on Covertype data. The labels comprehensibly indicate how different methods trade certain efficiency properties via color-coded ratings. }
    \label{fig:labels}
\end{figure}

\begin{figure}
\centering
    \includegraphics[width=.495\linewidth]{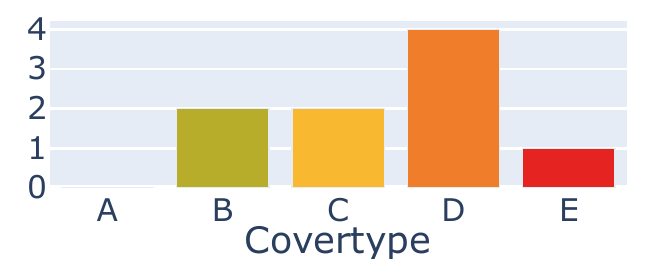}
    \hfill
    \includegraphics[width=.495\linewidth]{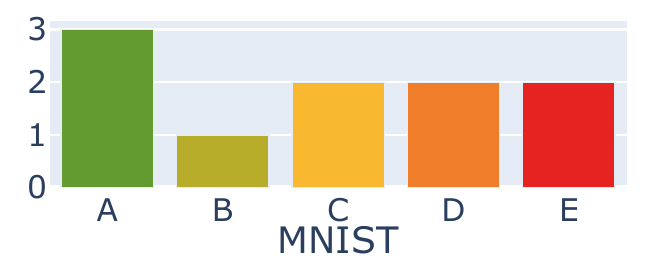}
    \\
    \includegraphics[width=.495\linewidth]{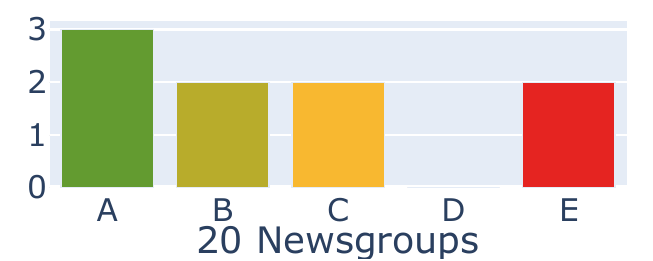}
    \hfill
    \includegraphics[width=.495\linewidth]{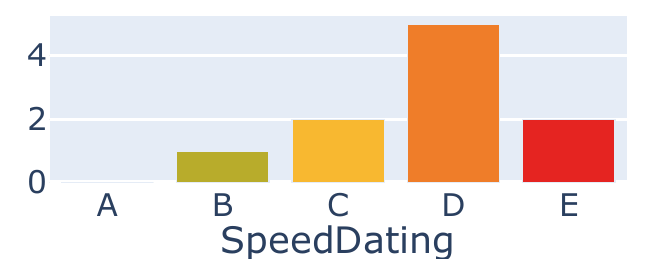}
    \\
    \includegraphics[width=.495\linewidth]{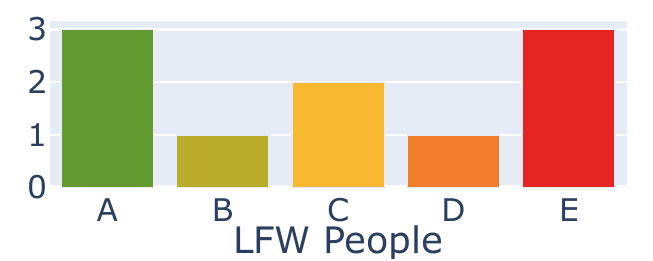}
    \hfill
    \includegraphics[width=.495\linewidth]{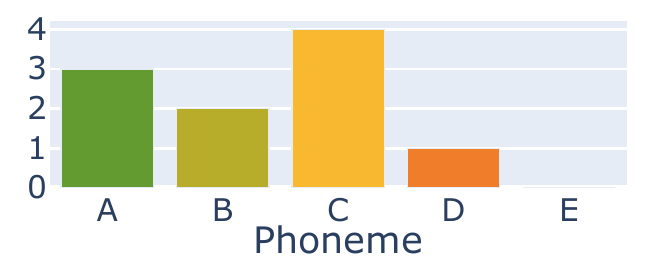}
    \\
    \caption{Distributions of method efficiency for the biggest data sets. The proposed method of assessing efficiency provides useful information for all applications. }
    \label{fig:dataset_distribs}
\end{figure}

Let us take a closer look and compare the different methods on this benchmark.
Figure \ref{fig:scatter-tradeoff} depicts a scatter plot with the model positioning in the power draw versus $F_1$ score trade-off.
As expected, the reference (RR) is placed at $(1, 1)$ coordinates.
The scatter plot is divided into colored grids which represent the discrete rating bins, while the point color indicates the compound rating.
In many cases those colors do not match, due to the plot only displaying two of the eight investigated metrics.
Note also that maximizing the index of power draw corresponds to minimizing the respective measured values thanks to $\sigma_i$ (cf. Equation \ref{eq:index}).
Energy labels for selected methods are depicted in Figure \ref{fig:labels}.
We see that while RF has slightly better accuracy scores, LR receives the same compound rating due to more resource-friendly computation.
For SVM however, the even better predictive quality cannot outweigh the immense computational effort.

Going back to the bigger picture, we can investigate the distributions of compound ratings as displayed in Figure \ref{fig:dataset_distribs}.
We can see that the proposed methodology indeed provides useful efficiency overviews:
Each data set has clear winning and loosing methods.
Note that these results were found with fixed rating bins across all metrics and applications, we merely chose different reference models for each data set.

\begin{figure}
\centering
    \includegraphics[width=.495\linewidth]{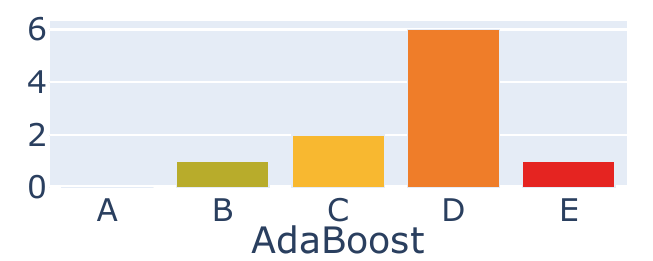}
    \hfill
    \includegraphics[width=.495\linewidth]{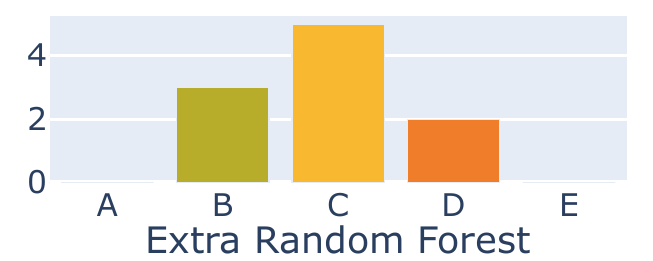}
    \\
    \includegraphics[width=.495\linewidth]{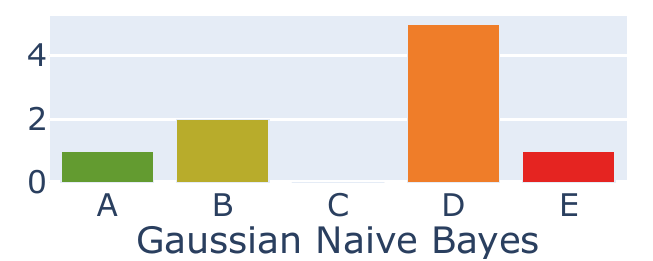}
    \hfill
    \includegraphics[width=.495\linewidth]{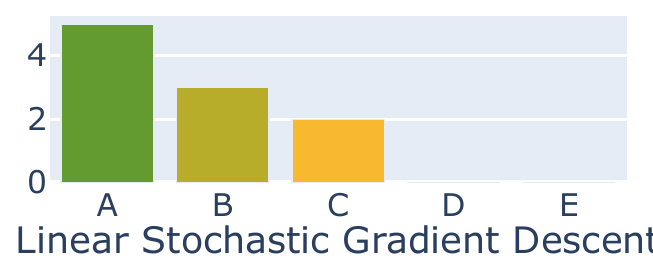}
    \\
    \includegraphics[width=.495\linewidth]{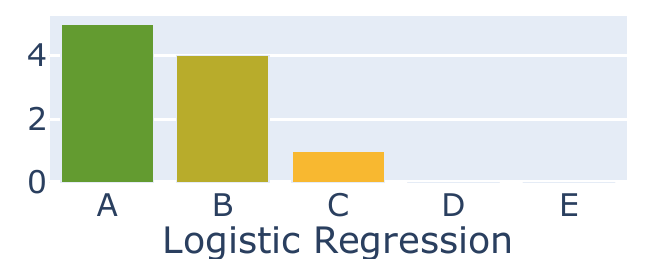}
    \hfill
    \includegraphics[width=.495\linewidth]{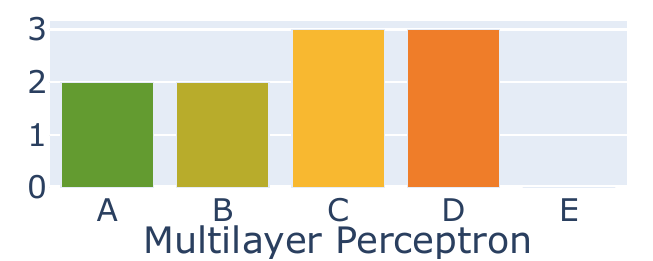}
    \\
    \includegraphics[width=.495\linewidth]{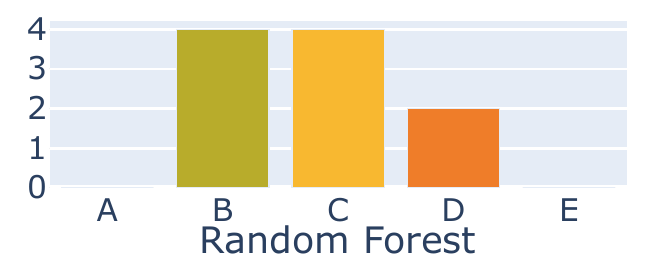}
    \hfill
    \includegraphics[width=.495\linewidth]{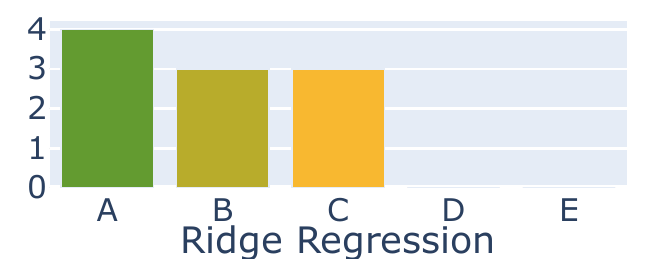}
    \\
    \includegraphics[width=.495\linewidth]{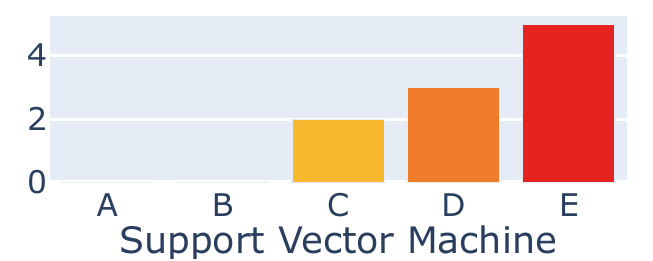}
    \hfill
    \includegraphics[width=.495\linewidth]{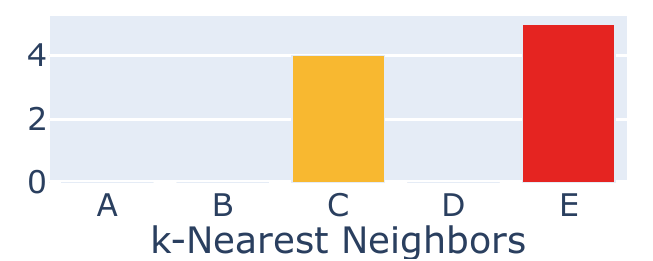}
    \caption{Distributions of compound ratings for each method over all ten data sets. In our empirical study, linear regression models seem to be most efficient.}
    \label{fig:model_distribs}
\end{figure}

We can also look at the distribution of ratings for each ML method across all tested data sets, as displayed in Figure \ref{fig:model_distribs}
In our empirical studies, some methods like linear regression variants seem to be more likely to perform efficiently.
We argue that such findings can be of great help for developers of AI systems, that want to find the most efficient model for their own application scenario.

\section{Conclusion}

In this work, we investigated the energy efficiency of deploying different ML methods on popular classification benchmarks.
For that we build upon and enhanced a recently published methodology \cite{sogood2022}.
As extensions, we added some metrics and reworked the weighting for assessing the compound rating of models.
The experiments illustrate how every data set has its unique landscape of efficiency, as every method trades certain aspects of performance.
Looking at statistics over all benchmarks, some methods are more likely to receive a high efficiency rating.
We hope that our findings shed some light onto the mostly undocumented efficiency trade-offs when choosing among ML methods.
We showed the feasibility of the proposed methodology, and hope to benefit developers in making the assemblage of future AI systems more resource-aware.

\section*{Acknowledgements}
This research has been funded by the Federal Ministry of Education and Research of Germany and the state of North-Rhine Westphalia as part of the Lamarr-Institute for Machine Learning and Artificial Intelligence.

\bibliography{aaai23}


\end{document}